\documentclass{article}
\usepackage{spconf,amsmath,graphicx}
\usepackage{subcaption}
\usepackage{hyperref}

\usepackage[english]{babel}
\usepackage[autostyle]{csquotes}

\usepackage{algorithm}
\usepackage{algorithmicx}
\usepackage{algpseudocode}

\usepackage[table]{xcolor}

\usepackage{booktabs}
\usepackage{multirow}
\usepackage{makecell}

\usepackage{siunitx}

\newcommand{\NA}{{\color{gray}N/A}}

\title{YawDD+: Frame-level Annotations for Accurate\\ Yawn Recognition on Edge Platforms}
\name{Ahmed Mujtaba\(^1\), Gleb Radchenko\(^1\), Marc Masana\(^2\), and Radu Prodan\(^3\)
\thanks{This paper is accepted in the 33rd IEEE International Conference on Image Processing (ICIP) 2026.}
}
\address{\(^1\)Embedded Systems Division, Silicon Austria Labs, Graz, Austria\\
\(^2\)Institute of Visual Computing, Graz University of Technology, Austria\\
\(^3\)Department of Computer Science, University of Innsbruck, Austria}

\begin{document}
%
\maketitle
\begin{abstract}
Driver fatigue remains a leading cause of road accidents, responsible for 24\% of crashes. While yawning serves as an early behavioral indicator of fatigue, existing approaches face significant challenges due to the presence of systematic noise in video-annotated datasets arising from coarse temporal annotations. Training robust machine learning (ML) models requires rich supervisory labels that help learn salient features from the training data. Moreover, efficient on-device training and inference of models on edge devices is crucial in driver fatigue detection tasks to enable accurate real-time decisions on vehicles without reliance on cloud infrastructure. To address this issue, we develop a semi-automated labeling pipeline with human-in-the-loop verification to annotate YawDD videos to YawDD+ frame-level annotations, enabling more accurate model training on edge platforms such as NVIDIA Jetson NANO. Training the established MNasNet classifier and YOLOv11 detector architectures on YawDD+ improves frame accuracy by up to 6\% and mAP by 5\% over video-level supervision, achieving 99.34\% classification accuracy and 95.69\% detection mAP on Jetson NANO and AGX. Moreover, MNasNet completed the epoch time in just 8.69 min/epoch while delivering up to 115 frames-per-second (FPS) inference time on AGX, confirming that enhanced data quality alone supports on-device driver fatigue monitoring systems without server-side computation. The YawDD+ dataset and trained models are available online\footnote{\url{https://opensource.silicon-austria.com/mujtabaa/yawdd}}.
\end{abstract}
\section{Introduction}
\label{sec:intro}

Driver fatigue impairs alertness and reaction time, increasing the risk of road collisions. The US National Highway Traffic Safety Administration estimates that drowsiness leads to approximately \num{50000} injuries and nearly \num{800} deaths in 2017~\cite{nthsa}. Furthermore, approximately 24\% of car crashes involve fatigued or drowsy drivers~\cite{klauer2006impact}. Camera-sensors deployed on vehicles generate millions of data samples every day and ML models can be trained to detect drivers' fatigue by focusing on yawning, head pose, and blinking time. Among observable signs of fatigue, \enquote{yawning} has been widely studied as an important behavioral indicator of increased sleepiness and loss of alertness, making yawning an early warning sign of fatigue-related impairment.

Training ML models is conventionally performed on high-powered GPUs or cloud infrastructure; however, safety-critical applications, such as in-vehicle driver monitoring systems, demand edge-based training due to the following reasons: first, these systems must continuously adapt to unseen data distributions, including variations in drivers, environments, and lighting conditions, necessitating periodic on-device model updates; second, reliance on cloud infrastructure is often impractical due to limited or intermittent connectivity, particularly in remote or latency-sensitive scenarios; and third, on-device training preserves data privacy, as sensitive visual data (e.g., a driver's face) does not need to be transmitted to the cloud. On the other hand, existing fatigue datasets, such as YawDD~\cite{abtahi2014yawdd}, typically label entire videos as \enquote{yawning}, although most frames display unrelated actions, such as normal driving or conversation. Such video annotations introduce systematic noise into the training data by incorrectly associating frames exhibiting normal behavior with the \enquote{yawn} category. While some ML models, such as RNNs, LSTMs, and video transformers, can learn temporal dependencies and context within videos, they are expensive to train and deploy on constrained edge devices.

\begin{figure*}[t]
     \centering
     \includegraphics[width=0.85\textwidth, trim=0 4 0 6, clip]{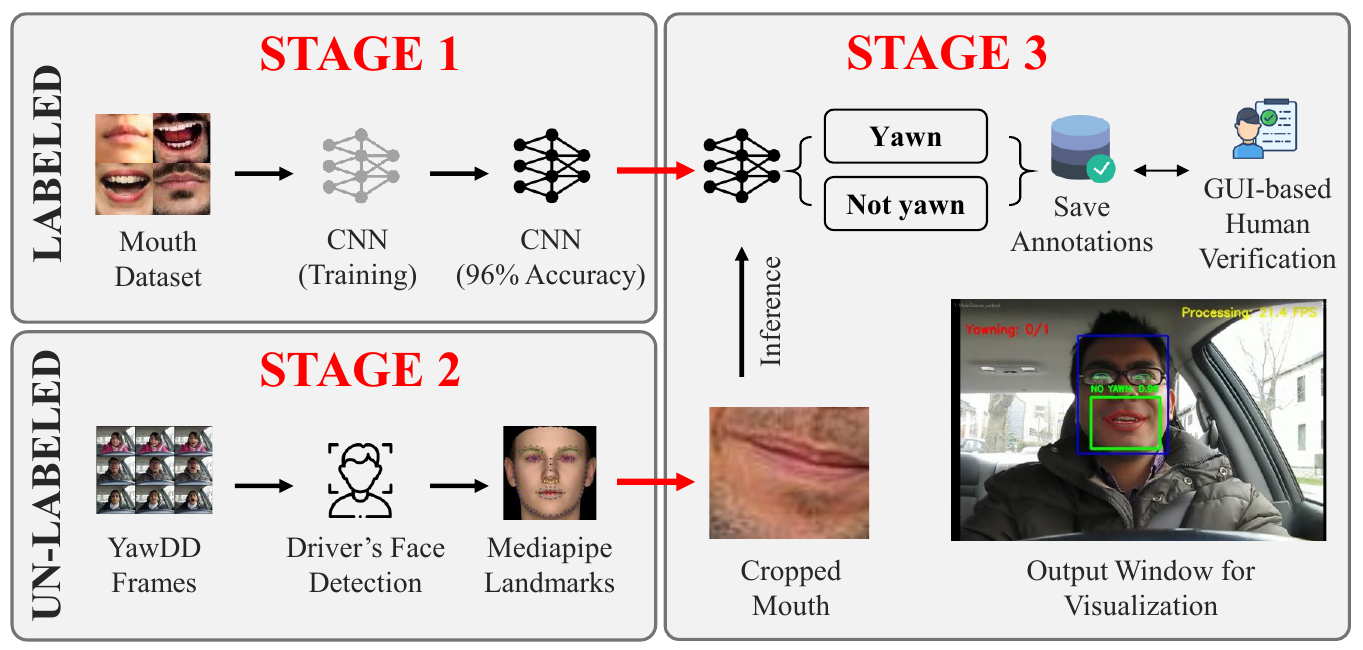}
     \caption[Semi]{YawDD+ semi-automated labeling pipeline with human-in-the-loop verification applied to YawDD~\cite{abtahi2014yawdd}.}
     \label{fig:labeling_pipeline}
\end{figure*}

In this work, we aim to eliminate label noise and improve model performance by building YawDD+, an annotated dataset that migrates the original YawDD video-based annotations to the frame level. We present a semi-automated pipeline that combines deep neural networks with human verification to correct label errors and annotate \num{124201} yawning and \num{222837} non-yawning images, generating precise frame-level annotations. We then perform on-device training and evaluation of \mbox{MNasNet}~\cite{mnasnet} for yawn classification and YOLOv11~\cite{khanam2024yolov11} for yawn detection on NVIDIA jetson devices using our refined YawDD+ annotations, which achieves higher accuracy than video-based baselines while fitting within the compute and memory limits of typical edge devices, removing the need for server-side processing and enabling practical in-vehicle driver fatigue monitoring.

\section{Related Works}
\label{sec:related_works}
YawDD~\cite{abtahi2014yawdd} comprehensively covers yawning patterns by capturing driver faces from different camera perspectives. However, this dataset contains video-based annotations with a substantial number of temporal frames that contain non-yawning features.

Bai et al.~\cite{bai2021two} proposed two-stream spatial–temporal graph convolutional networks (2s-STGCN) using video sequences that generate facial landmark detection results according to their spatial and temporal relationship, and achieved 93.4\% accuracy. Majeed et al.\cite{majeed2023detection} implement a hybrid CNN-RNN model to incorporate spatial-temporal features during training and deliver 95.64\% accuracy. Recently, DLS~\cite{xu2025novel} proposed a dual-lightweight swin-transformer model that incorporates Farneback optical flow to calculate the movement of pixels in video sequences to obtain time-dimensional features, and achieved 96.16\% accuracy. Civik et al.~\cite{civik2023real} developed a driver fatigue detection system to classify four different situations by analyzing the eye and mouth areas of the driver, and achieved 94.5\% accuracy with an overall 6 FPS on an NVIDIA Jetson NANO. Al-Mahbashi et al.~\cite{al2025real} developed a distracted driving detection system based on the improved YOLOv8, and attained 98.20\% accuracy with 43.17 FPS on NVIDIA Jetson Xavier NX (more powerful than NVIDIA Jetson NANO). Zhou and Li~\cite{zhou2021real} proposed a driver fatigue monitoring system based on MobileNetv3 and achieved 94\% accuracy with 22 FPS on NVIDIA Jetson TX2. He et al.~\cite{he2020real} used a two-staged CNN on YawDD, which includes a detection followed by a classification phase designed to extract facial features and localize the eyes and mouth regions with 93.83\% accuracy and 96.3ms inference time on a Raspberry Pi 4.

All of the studies above rely on YawDD video-level annotations and do not evaluate on edge platforms. On-device training is critical for automotive applications where private information cannot be shared with cloud, and must be computed real-time, generally below 100\,ms end-to-end, to avoid safety-critical delays in decision-making~\cite{shi2022vips}.

\section{Semi-Automated Pipeline\\for Labeling Yawn Datasets}
\label{sec:labeling_pipeline}
YawDD~\cite{abtahi2014yawdd} provides two in-vehicle camera views: a)~\textit{Dashboard} includes 29 videos, one per subject, each covering silent driving, conversational driving, and yawning episodes; and b)~\textit{Rear-view} contains 322 videos grouped into three behavioral states: normal driving, talking or singing while driving, and yawning while driving.
Each participant contributes to three or four sequences, producing approximately \num{287225} frames for precise annotation. Manual annotation of this large corpus requires significant hours and careful effort to avoid mislabels that could impair model generalization.

We propose a semi-automated labeling pipeline that employs ML models for intelligent annotation assistance, consisting of three interconnected stages, as illustrated in Figure~\ref{fig:labeling_pipeline}, and detailed in the remainder of this section.

\begin{figure}[t]
     \centering
     \includegraphics[width=0.9\columnwidth, trim={0cm, 5cm, 6.2cm, 0cm}, clip]{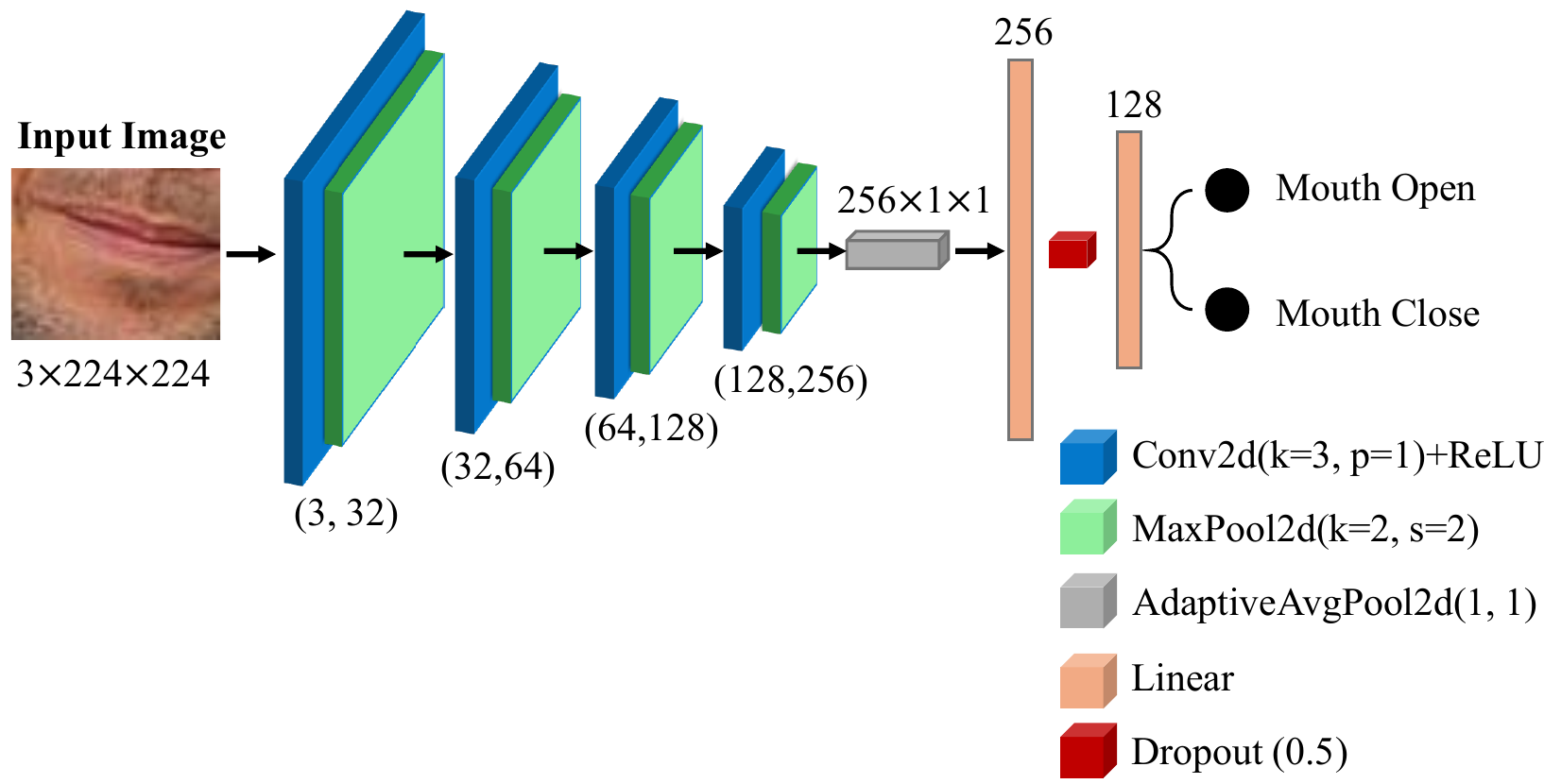}
     \caption{Mouth state binary classification model architecture.}
     \label{fig:mouth_network}
\end{figure}

\subsection{Stage~1: Mouth state classification}
Finding suitable datasets for mouth-state classification proved challenging at this stage. Existing publicly available datasets were limited, with the most comprehensive dataset~\cite{yawn-dataset} containing only \num{5119} images encompassing both color and grayscale representations with varying spatial resolutions. To overcome the limited dataset size, we implemented several data augmentation strategies including rotation, scaling, brightness adjustment, and horizontal flipping. We then train a CNN architecture for binary classification of mouth states into \enquote{open} and \enquote{close}. Figure~\ref{fig:mouth_network} shows the shallow, lightweight architecture used for training containing approximately \num{421000} trainable parameters, achieving 96\% test accuracy.

\subsection{Stage~2: Face detection and landmarks}
We extract \num{287225} frames containing yawning and non-yawning patterns for comprehensive annotation. The annotation process requires two critical steps: (1) driver face detection, and (2) precise mouth region extraction to apply the trained CNN model from Stage~1.

We employ YOLOv8~\cite{varghese2024yolov8} for robust face localization across diverse lighting conditions and poses characteristic of vehicle's driving environments. The detector operates with configurable confidence thresholds and implements non-maximum suppression to handle multiple faces within a single frame, due to passengers, reflections, or background individuals. Precise driver identification was critical for maintaining annotation accuracy. Therefore, we implement an area-based face selection method that computes the spatial area of each detected bounding box and automatically selects the face with maximum area. This heuristic proved highly effective as driver faces consistently occupied the largest spatial region in dashboard and rear-view mounted camera perspectives.

For precise localization of the mouth region, we integrate MediaPipe Face Mesh~\cite{mediapipe} to extract \num{468} three-dimensional facial landmarks with sub-pixel precision. The Dlib facial landmark detector~\cite{dlib} exhibited significant temporal instability and imprecise localization around the mouth, particularly under varying illumination and head pose conditions common in vehicular environments. The mouth region extraction algorithm utilizes a comprehensive set of lip-specific landmarks including upper lip (indices: 61, 185, 40, 39, 37, 267, 269, 270, 409, 291), lower lip (indices: 0, 17, 314, 405, 321, 375, 89, 87, 14, 317, 402, 318), and contextual surrounding landmarks (indices: 78, 191, 80, 81, 82, 313, 312, 311, 310, 415, 308) to ensure complete coverage, with configurable padding expansion (10 pixels) to accommodate mouth movement variations and ensure complete lip visibility for better classification accuracy.

\begin{figure}[t]
    \centering
    \includegraphics[width=0.8\columnwidth, trim=0 10 0 10, clip]{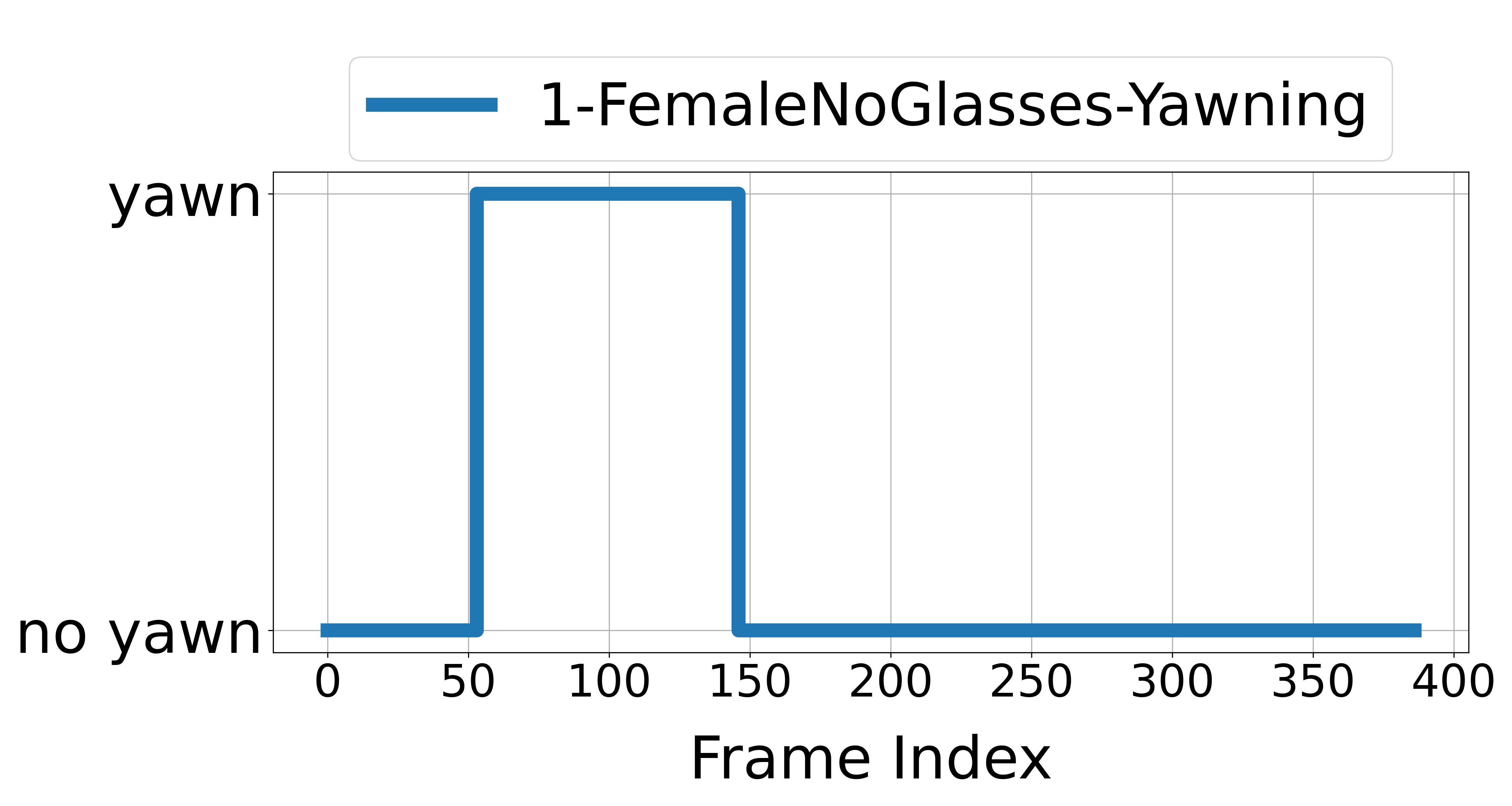}\\
    \includegraphics[width=0.8\columnwidth, trim=0 10 0 10, clip]{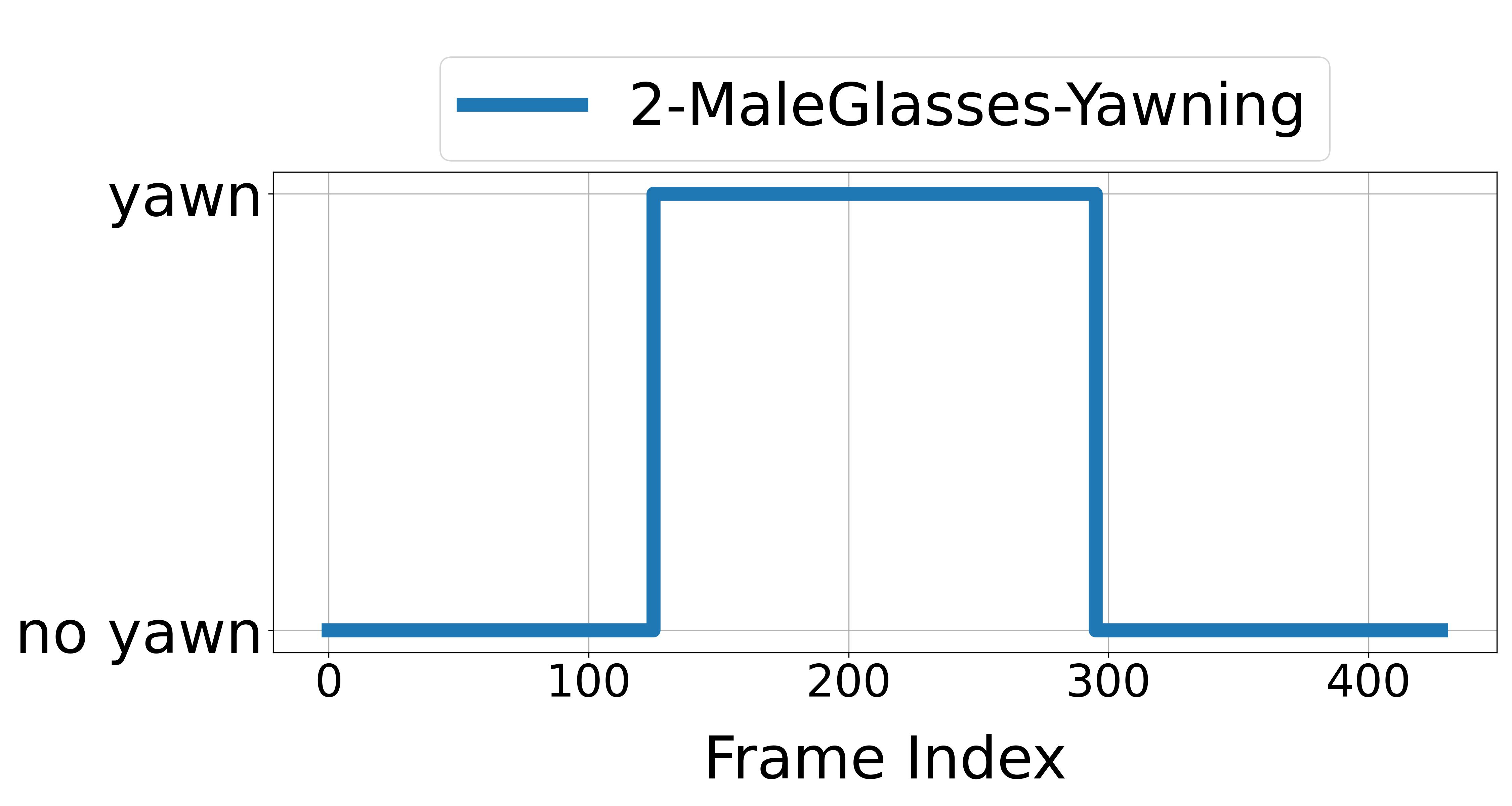}
    \caption{Comparison of fatigued driver yawning episodes. Subject \texttt{1-FemaleNoGlasses} yawns for 93/387 frames, while \texttt{2-MaleGlasses} yawns for 170/429 frames. Remaining frames are mislabeled by YawDD annotations and contain non-yawn behaviors (talking, smiling, or neutral).}
    \label{fig:annotation_analysis}
\end{figure}

\begin{table*}[t]
\centering
\caption{Comparative yawning classification (Cls) and detection (Det) results using YawDD and YawDD+ annotations.}
\label{tab:comparison}
\resizebox{\textwidth}{!}{%
\begin{tabular}{@{}lccccccc@{}}
\toprule
\textit{Approach} & \textit{Method} & \textit{Annotation} & \textit{Task} & \textit{Accuracy\,/\,mAP} & \textit{Edge Device} & \textit{Inference} & \textit{Train (epoch)} \\ \midrule
\multirow{3}{*}{Video-based} & 2s-STGCN~\cite{bai2021two} & YawDD & Cls & 93.4\% & \NA & \NA & \NA  \\
& CNN-RNN~\cite{majeed2023detection} & YawDD & Cls  & 96.6\% & \NA & \NA   & \NA \\
& DLS~\cite{xu2025novel} & YawDD & Cls  & 96.14\% & \NA & \NA & \NA \\ \midrule
\multirow{8}{*}{Frame-based} & Two-stage CNN~\cite{he2020real} & YawDD & Cls & 93.83\% & Raspberry Pi 4  & 96.3 ms & \NA \\
& CNN~\cite{civik2023real} & YawDD & Cls & 94.5\% & Jetson NANO & 166 ms & \NA \\
& CNN~\cite{essahraui2025real} & YawDD & Cls & 93.31\% & \NA & \NA & \NA  \\
& YOLOv5, YOLOv8~\cite{essahraui2025real} & YawDD & Det & 90.1\% mAP, 90.3\% mAP   & \NA & \NA & \NA  \\
& GM-YOLOv8n~\cite{al2025real} & YawDD & Det & 98.20\% mAP   & Jetson Xavier NX & 23.16 ms & \NA  \\
& MobileNetv3~\cite{zhou2021real} & YawDD & Cls & 94\% & Jetson Xavier TX2 & 45.4 ms & \NA  \\ \cmidrule(l){2-8}
& \multirow{2}{*}{MNasNet, YOLOv11} & \multirow{2}{*}{YawDD+ (Ours)} & \multirow{2}{*}{Cls, Det} & \multirow{2}{*}{99.34\%, 95.69\% mAP} & Jetson NANO & 16.71 ms, 35.7 ms & 13.93 min, 85.8 min \\
& & & & & Jetson AGX & 12.03 ms, 24.33 ms & 8.69 min, 41.4 min \\ \bottomrule
\end{tabular}}
\end{table*}

\subsection{Stage~3: Automated annotations}
Finally, we incorporate the human-in-the-loop verification in this automated labeling pipeline. The cropped mouth regions (from Stage~2) serve as input to the already trained CNN classifier from Stage~1, which outputs binary predictions and confidence scores. Predictions with high confidence score ($>95\%$) are used to label the images as \enquote{yawn}, while the low confidence images are verified using the human-in-the-loop GUI. The GUI contains a custom interface that loads batches of 64 images and their automated predictions, enabling real-time error correction of false positives and false negatives. An empirical evaluation of automated labeling accuracy shows that approximately 80\% of annotations are correct, substantially reducing man-hours while maintaining high annotation quality. The remaining 20\% of cases requiring manual human intervention correspond to challenging edge cases, such as extreme lighting conditions or ambiguous mouth positions. As a result, human effort is focused on targeted verification and correction rather than exhaustive annotation, making the pipeline considerably less time-consuming in practice. The validated annotations are automatically linked with the corresponding images, and the resulting dataset is named as YawDD+ stored in structured formats compatible with standard ML frameworks.

\section{Experimental Results}
\label{sec:experiments}
We conducted various experiments to demonstrate the effectiveness of YawDD+ annotations. Section \ref{subsec:annotation_analysis} differentiates the annotation quality of YawDD and YawDD+. Section \ref{subsec:results_and_comparison} provides our state-of-the-art results and on-device training on edge devices as well as compares with existing YawDD baselines studies.

\subsection{Annotation Analysis}
\label{subsec:annotation_analysis}
We annotated a total of \num{287225} frames from different camera perspectives (dashboard and rear-view) using our semi-automated labeling pipeline. The annotation taxonomy classifies \enquote{normal} and \enquote{talking} behaviors under the \enquote{no-yawn} category, while labeling yawning instances under the \enquote{yawn} category. The labeling analysis reveals a significant class imbalance, with \num{22875} frames containing yawning behavior and \num{264350} frames representing \enquote{no-yawn} instances. Figure~\ref{fig:annotation_analysis} illustrates the comparison of YawDD and YawDD+ annotations. The YawDD annotations classifies the whole video of fatigued subject as \enquote{yawning}, whereas YawDD+ annotations accurately identifies the yawn patterns at the frame-level. For instance, \texttt{1-FemaleNoGlasses} yawns in \num{93} consecutive frames while the remaining frames involves non-yawn behavior. These driver-specific behaviors present considerable heterogeneity across individual subjects, underscoring the challenge of developing generalized models that can effectively adapt to diverse individual behavioral patterns, while maintaining consistent performance across different drivers and driving contexts.

\begin{table}[t]
\caption{Accuracy and inference time of transformer-based architectures on high-end (A100) and edge-based (AGX, NANO) NVIDIA GPUs using YawDD+.}
\label{tab:temporal}
\resizebox{\columnwidth}{!}{%
\begin{tabular}{@{}rlcccccc@{}}
\toprule
\multirow{2}{*}{\textbf{Models}} & \multirow{2}{*}{\textbf{(size)}} & \multirow{2}{*}{\textbf{\begin{tabular}[c]{@{}c@{}}Clip\\ Frames\end{tabular}}} & \multirow{2}{*}{\textbf{\begin{tabular}[c]{@{}c@{}}Accuracy\\ (\%)\end{tabular}}} & \multirow{2}{*}{\textbf{\begin{tabular}[c]{@{}c@{}}F1-Score\\ (\%)\end{tabular}}} & \multicolumn{3}{c}{\textbf{Inference (sec)}} \\ \cmidrule(l){6-8}
 & & & & & \textbf{A100} & \textbf{AGX} & \textbf{NANO} \\ \midrule
\multirow{3}{*}{\texttt{mc3\_18}} & \multirow{3}{*}{(131.6 MB)} & 16 & 80.77 & 78.91 & 0.57 & 0.49 & 3.68 \\
 & & 32 & 84.62 & 83.12 & 0.86 & 0.72 & 4.45 \\
 & & 64 & 80.77 & 79.23 & 1.2  & 1.14 & 4.77 \\ \midrule
\multirow{3}{*}{\texttt{r3d\_18}} & \multirow{3}{*}{(379.7 MB)} & 16 & 69.23 & 68.88 & 0.48 & 0.45 & 3.26 \\
 & & 32 & 76.92 & 75.06 & 0.89 & 0.69 & 3.73 \\
 & & 64 & 80.77 & 78.66 & 1.2  & 1.15 & 4.62 \\ \midrule
\multirow{3}{*}{\texttt{r2plus1d\_18}} & \multirow{3}{*}{(358.5 MB)} & 16 & 76.92 & 75.10 & 0.64 & 0.69 & 3.51 \\
 & & 32 & 84.62 & 82.78 & 0.89 & 0.94 & 3.93 \\
 & & 64 & 84.62 & 82.78 & 1.04 & 1.41 & 4.88 \\ \midrule
\multirow{3}{*}{\texttt{swin3d\_t}} & \multirow{3}{*}{(333 MB)} & 16 & 84.62 & 82.81 & 0.5  & 1.02 & 3.84 \\
 & & 32 & 88.46 & 86.65 & 0.67 & 1.40 & 4.89 \\
 & & 64 & 88.46 & 86.65 & 1.003 & 2.08 & 6.81 \\ \midrule
\multirow{3}{*}{\texttt{swin3d\_s}} & \multirow{3}{*}{(595.2 MB)} & 16 & 84.62 & 82.81 & 0.47 & 1.14 & 4.37 \\
 & & 32 & 88.46 & 86.65 & 0.68 & 1.66 & 5.84 \\
 & & 64 & 92.31 & 90.47 & 0.91 & 2.29 & 8.56 \\ \midrule
\multirow{3}{*}{\texttt{swin3d\_b}} & \multirow{3}{*}{(1 GB)} & 16 & 88.46 & 86.65 & 0.37 & 1.23 & 4.99 \\
 & & 32 & 88.46 & 86.65 & 0.57 & 1.82 & 7.02 \\
 & & 64 & 92.31 & 90.47 & 1.13 & 2.57 & 10.79 \\ \bottomrule
\end{tabular}}
\end{table}

\subsection{Results \& Comparison}
\label{subsec:results_and_comparison}
\textit{Experimental Setup}: For training and inference on edge devices, we utilized NVIDIA Jetson Orin NANO 8GB (6-core 64-bit ARM Cortex-A78AE, and Orin AGX 64GB (12-core 64-bit ARM Cortex-A78AE). For fair evaluation, we applied YawDD+ to existing models (MNasNet~\cite{mnasnet}, YOLOv11~\cite{khanam2024yolov11}) to highlight the robustness of YawDD+ annotations. The video frames from each subject were pooled together, randomly shuffled with a fixed random seed (42), split into an 80/20 train/validation ratio. The comparative evaluation of our results are summarized in Table~\ref{tab:comparison}. 

Video-based approaches exploit temporal dependencies using specialized transformer and RNN-based architectures such as CNN–RNN hybrids~\cite{majeed2023detection}, 2s-STGCN~\cite{bai2021two}, or DLS~\cite{xu2025novel}. While these architectures achieve competitive classification accuracy (up to 96.6\%), they rely on video-level annotations and do not report inference latency or training times on edge devices, making them less suitable for real-time deployment on embedded or edge-based platforms. Table~\ref{tab:temporal} reports the high inference cost of transformer-based architectures on edge hardware. In contrast, frame-based approaches focus on per-frame learning using lightweight convolutional architectures, enabling efficient inference and easier integration into real-time systems. Prior frame-based studies~\cite{he2020real, civik2023real, essahraui2025real, al2025real, zhou2021real} trained on YawDD video-level annotations report classification accuracies ranging from 93.31\% to 94.5\% and detection performance up to 98.20\% mAP. However, they omit important efficiency metrics, such as training costs on edge devices, which are critical for evaluating their real-world applicability.

The models trained on YawDD+ achieve superior performance compared to prior work. Specifically, MNasNet achieves a classification accuracy of 99.34\% and 98.30\% F1-score , outperforming both state-of-the-art frame-based classifiers and more computationally demanding transformer architectures (see Table~\ref{tab:temporal}). Similarly, YOLOv11 attains 95.69\% mAP$_{50\text{–}95}$ and 98.67\% F1-score, significantly exceeding the detection performance of YOLOv5~\cite{essahraui2025real} and YOLOv8~\cite{essahraui2025real} while maintaining substantially lower inference latency. Importantly, our evaluation includes deployment on multiple edge platforms, namely Jetson NANO and Jetson AGX, demonstrating consistent real-time performance with inference times as low as 12.03 ms for classification and 24.33 ms for detection. In addition to inference efficiency, we also report per-epoch training time, which is largely absent in prior work. The relatively short training durations of MNasNet and YOLOv11 further highlight their real-world applicability for edge training and deployment. 

\begin{figure}[t]
    \centering
    \begin{subfigure}[b]{0.53\columnwidth}
        \centering
        \includegraphics[width=\linewidth, trim={1cm, 1cm, 1cm, 3cm}, clip]{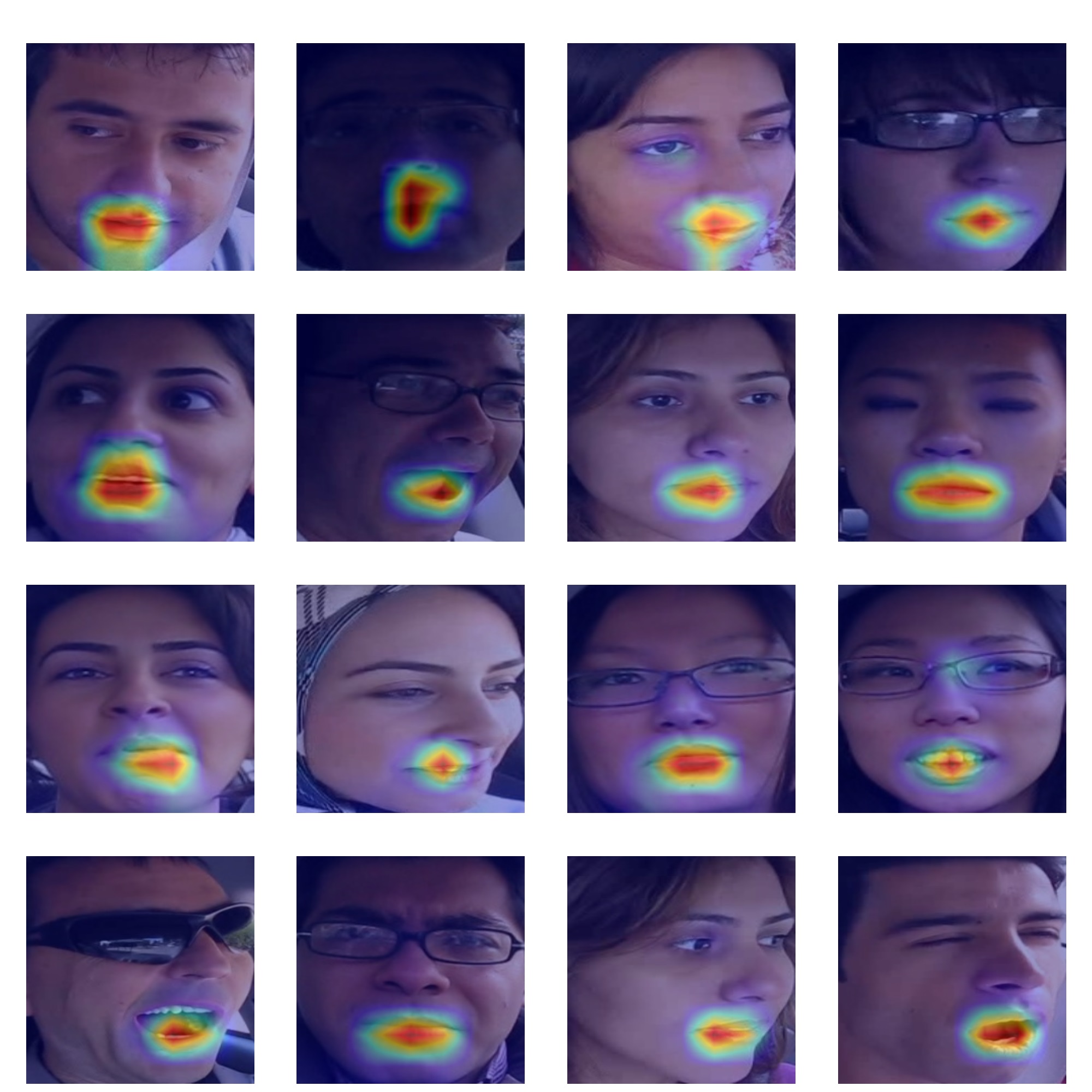}
        \caption{Grad-CAM visualization of MNasNet predictions.}
        \label{fig:gradcam}
    \end{subfigure}
    \hfill
    \begin{subfigure}[b]{0.44\columnwidth}
        \centering
        \includegraphics[width=\linewidth]{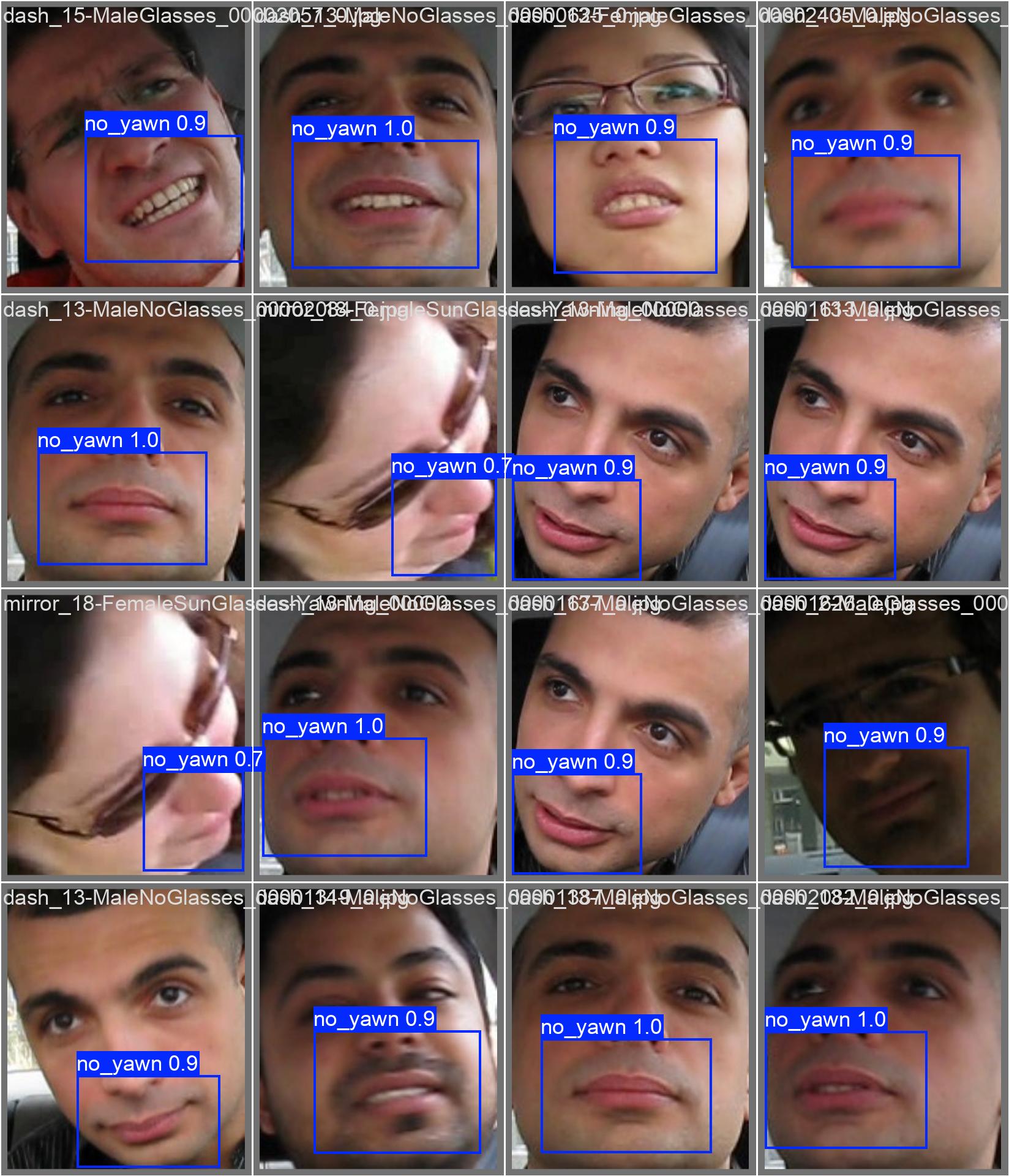}
        \caption{Mouth localization with YOLOv11.}
        \label{fig:yolov11}
    \end{subfigure}
    \caption{Qualitative results using YawDD+ annotations: (a) highlights salient regions learned by MNasNet via Grad-CAM, while (b) shows accurate mouth localization using YOLOv11.}
    \label{fig:qualitative}
\end{figure}


\section{Discussion \& Analysis}
Distinguishing yawning from visually similar facial behaviors such as talking and smiling is a complex task owing to dynamic mouth movements, partial or full mouth opening, and overlapping lip and facial muscle configurations. These ambiguities are further exacerbated in real-world driving environments, where illumination changes, head pose variations, occlusions (e.g., hands or sunglasses), and camera-perspective distortions are common. As shown in Figure~\ref{fig:annotation_analysis}, video-level annotations in YawDD incorrectly associate many non-yawning frames with the “yawn” label, introducing systematic noise that hinders robust model learning. YawDD+ mitigates this issue by providing precise frame-level annotations that offer rich supervisory signals with well-defined, discriminative visual patterns, enabling effective learning and generalization of yawning-related facial dynamics. The high accuracy observed across the evaluated architectures (MNasNet and YOLOv11) is primarily thanks to the frame-based YawDD+ annotations. We employ gradient-weighted class activation mapping (\mbox{Grad-CAM}~\cite{selvaraju2017grad}) to visualize the spatial regions and salient features within the driver’s facial area that contribute most to the model’s predictions (see Fig.~\ref{fig:gradcam}). MNasNet consistently highlights the mouth region across varying illumination conditions as the most relevant for its decisions. Furthermore, YOLOv11n shows accurate spatial localization around the mouth region (see Fig.~\ref{fig:yolov11}), corroborating our findings and confirming the model’s capability to jointly learn discriminative features and precise region-of-interest localization for yawning detection.

Beyond automotive driver monitoring systems, YawDD+ can serve in related domains that require fine-grained analysis of facial behavior, including human–machine interaction, occupational safety monitoring, and assistive technologies for alertness assessment. The dataset’s ability to distinguish frame-based yawning from other facial movements also makes it valuable for multimodal fatigue analysis, where yawning is used among other behavioral indicators, such as eye closure or head pose. Additionally, the semi-automated pipeline (Figure~\ref{fig:labeling_pipeline}) can be migrated to other ML datasets to reduce manual labeling hours involved in vision-related tasks. 


\section{Conclusion}
\label{sec:conclusion}
This paper introduced a semi-automated labeling pipeline that upgrades YawDD to YawDD+, delivering precise frame-level annotations and removing label noise. With these annotations, MNasNet reaches 99.34\% accuracy, and YOLOv11 achieves 95.69\% mAP without any optimization, surpassing existing frame-based and video-based methods in accuracy by 3–6\%, while operating at 28–115 FPS on commodity edge hardware, compared to 6–10 FPS in prior work. Overall, the experimental findings validate that high-quality frame-level annotations in YawDD+ enable lightweight architectures to achieve superior accuracy, on-device training, and real-time inference, making them particularly well-suited for embedded driver monitoring systems in intelligent transportation and automotive safety applications. Future work will explore quantization, pruning, and distillation to further improve performance. We will also investigate applications such as federated learning to enable privacy-preserving collaborative model updates across distributed vehicles.

\section*{Acknowledgment}
This work received funding from the European Union MSCA-COFUND project CRYSTALLINE (grant num. 101126571, and the Austrian Research Promotion Agency (FFG grant agreement 909989 ``AIM AT Stiftungsprofessur für Edge AI''). This work has been supported by Silicon Austria Labs (SAL) owned by the Republic of Austria, the Styrian Business Promotion Agency (SFG), the federal state of Carinthia, the Upper Austrian Research (UAR), and the Austrian Association for the Electric and Electronics Industry (FEEI). This research was partially funded by the Austrian Science Fund (FWF) 10.55776/COE12.

\bibliographystyle{IEEEtran}
\bibliography{refs}

\end{document}